\documentclass[journal]{IEEEtran}
\usepackage[table]{xcolor}
\usepackage{colortbl}
\usepackage{amsmath,amsfonts}
\usepackage{graphicx}
\usepackage{amsmath}
\usepackage{amssymb}
\usepackage{booktabs}

\pdfoutput=1
\usepackage{algorithm}
\usepackage{array}
\usepackage[caption=false,font=normalsize,labelfont=sf,textfont=sf]{subfig}
\usepackage{textcomp}
\usepackage{stfloats}
\usepackage{url}
\usepackage{verbatim}
\usepackage{graphicx}
\usepackage{subfig}
\usepackage{cite}
\usepackage{float}
\usepackage[export]{adjustbox} 
\usepackage{siunitx}
\usepackage{multirow}
\usepackage{threeparttable}
\usepackage{color}
\usepackage{amsmath, bm}
\usepackage{bbding}
\usepackage[backref]{hyperref}
\usepackage{pifont}
\usepackage[all]{hypcap}
\usepackage{makecell}
\usepackage{lineno}
\usepackage{arydshln}
\usepackage{amsmath} 
\usepackage{amssymb}
\usepackage{xcolor}
\usepackage{mathtools}
\usepackage{multirow}

\usepackage{algorithm}
\usepackage{booktabs}
\usepackage{cuted}   
\usepackage{cleveref}
\usepackage{subcaption}
\usepackage{adjustbox}
\usepackage{algorithm}
\usepackage{algpseudocode}
\usepackage{lipsum}
\usepackage{array}
\newcolumntype{M}[1]{>{\centering\arraybackslash}m{#1}}  

\usepackage{array} 
\usepackage{booktabs} 
\usepackage{amsmath} 
\usepackage{multirow} 

\usepackage[dvipsnames]{xcolor} 

\begin{document}
\title{HiFi-MambaV2: Hierarchical Shared-Routed MoE for High-Fidelity MRI Reconstruction}

\author{\IEEEauthorblockN{Pengcheng Fang\IEEEauthorrefmark{1},
                        Hongli Chen\IEEEauthorrefmark{1},
                         Guangzhen Yao, 
                         Jian Shi, 
                         Fangfang Tang, 
                         Xiaohao Cai, 
                         Shanshan Shan\IEEEauthorrefmark{2},
                         Feng Liu 
                        }
                        
\thanks{\IEEEauthorrefmark{1} Equal Contribution.
\\
\IEEEauthorrefmark{2} Corresponding author.
}}

\markboth{Journal of \LaTeX\ Class Files,~Vol.~14, No.~8, August~2021}%
{Shell \MakeLowercase{\textit{et al.}}: A Sample Article Using IEEEtran.cls for IEEE Journals}

\maketitle

\begin{abstract}
Reconstructing high-fidelity MR images from undersampled k-space data requires recovering high-frequency details while maintaining anatomical coherence. We present HiFi-MambaV2, a hierarchical shared-routed Mixture-of-Experts (MoE) Mamba architecture that couples frequency decomposition with content-adaptive computation. The model comprises two core components: (i) a separable frequency-consistent Laplacian pyramid (SF-Lap) that delivers alias-resistant, stable low- and high-frequency streams; and (ii) a hierarchical shared-routed MoE that performs per-pixel top-1 sparse dispatch to shared experts and local routers, enabling effective specialization with stable cross-depth behavior. A lightweight global context path is fused into an unrolled, data-consistency-regularized backbone to reinforce long-range reasoning and preserve anatomical coherence. Evaluated on fastMRI, CC359, ACDC, M4Raw, and Prostate158, HiFi-MambaV2 consistently outperforms CNN-, Transformer-, and prior Mamba-based baselines in PSNR, SSIM, and NMSE across single- and multi-coil settings and multiple acceleration factors, consistently surpassing consistent improvements in high-frequency detail and overall structural fidelity. These results demonstrate that HiFi-MambaV2 enables reliable and robust MRI reconstruction.
\end{abstract}    

\vspace{-0.05in}
\section{Introduction}

\begin{figure}[tbp]
\centering
\includegraphics[width=1\columnwidth]{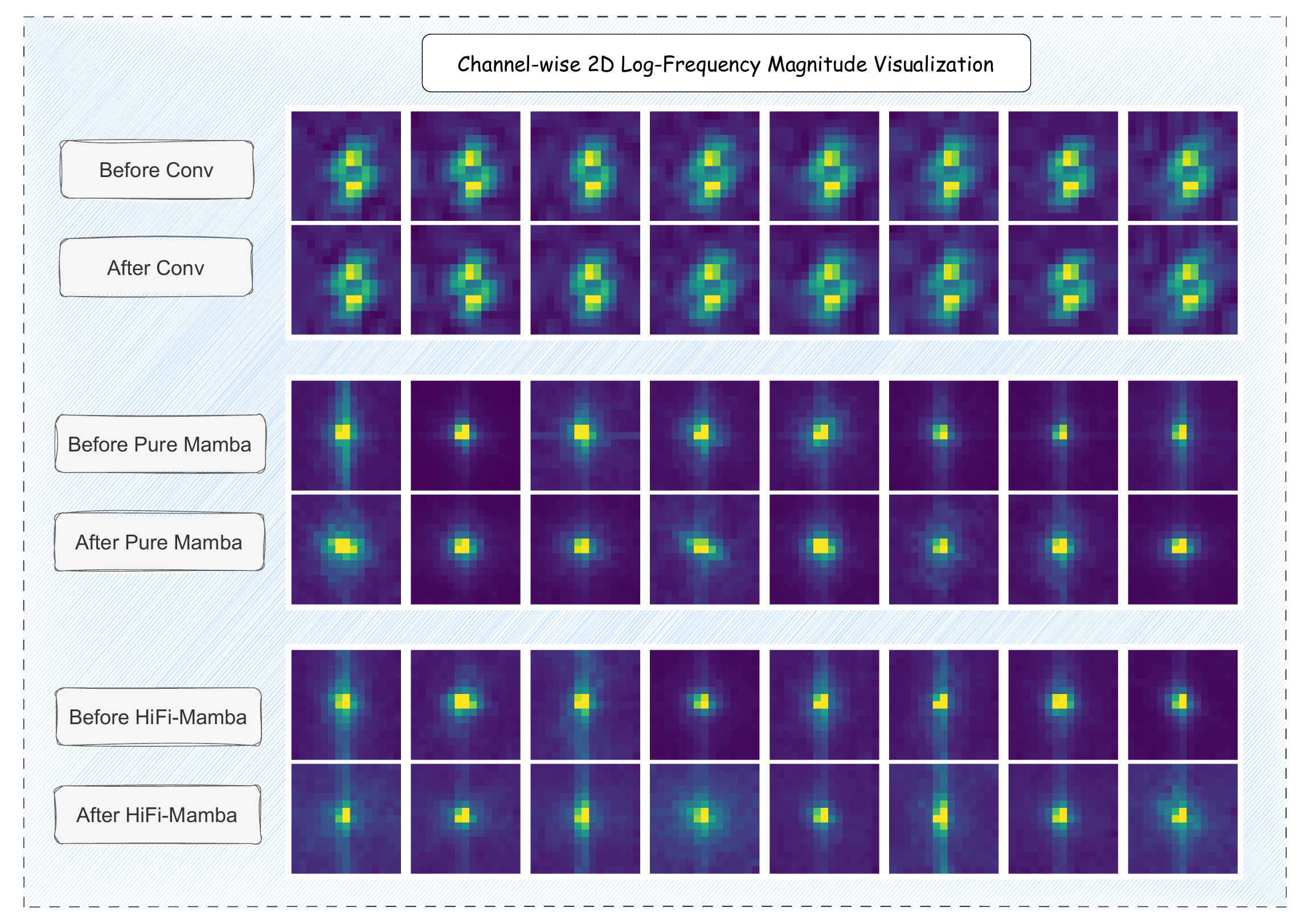}
\caption{Each panel visualizes channel-wise 2D log-magnitude Fourier spectra (14×14 thumbnails) of intermediate feature maps; top row is before the block and bottom row is after. 
Conv Block preserves pronounced cross-shaped high-frequency responses, reflecting strong sensitivity to local edges and textures. Pure Mamba suppresses mid/high frequencies and concentrates energy near the DC component, evidencing a low-pass, global-context behavior. Our HiFi–Mamba block produces a balanced spectrum: low frequencies dominate (global consistency) while moderate high-frequency energy is retained (local detail). This frequency-domain complementarity explains the downstream gains we report—integrating Mamba with convolution unifies global semantics and local precision.}
\label{fig1}
\end{figure}

Magnetic resonance imaging (MRI) is clinically indispensable for its non-invasive nature and superior soft-tissue contrast, but its long acquisition time increases patient discomfort and constrains clinical throughput~\cite{zbontar2018fastmri,shan2024b0inhomogeneity}. Performing reconstruction from undersampled k-space is an acceleration strategy, yet undersampling introduces aliasing artifacts~\cite{ye2019compressed}. This necessitates reconstruction methods that recover fine high-frequency details while preserving global anatomical coherence.

Deep learning has advanced MRI reconstruction by learning data-driven priors from large-scale datasets. Early works primarily utilized convolutional neural networks (CNNs) as the backbone ~\cite{schlemper2017deep, wang2016accelerating, qin2019crnn,hammernik2018variational}. However, CNNs are limited in modeling long-range dependencies, which are essential for preserving global anatomical consistency. Vision Transformers ~\cite{dosovitskiy2020image} later gained popularity for undersampled MRI reconstruction due to their ability to model long-range dependencies, but the quadratic complexity of self-attention hinders scaling to high-resolution clinical inputs and large fields of view~\cite{you2021unsupervised, wang2023dual}.

More recently, the selective state-space model (Mamba)~\cite{gu2023mamba} provides linear-time long-range modeling via input-dependent state transitions and gating, and has been adapted to vision tasks. However, direct application to MRI remains challenged by (i) redundant multi-directional scanning, (ii) local state parameterization that weakens anatomical coherence, and (iii) the lack of explicit frequency control. Our previous \textbf{HiFi-Mamba}~\cite{chen2025hifi} adopted an undirectional scanning dual-stream Mamba design. While effective, strict decoupling reduced cross-frequency interaction and global aggregation. In addition, static routing with a $w$-Laplacian transform offered limited frequency adaptivity (see the motivation in Section \ref{sec:motivation}).

To further analyze these behaviors, we visualize in Figure~\ref{fig1} the frequency-domain spectra of intermediate features before and after each block. Conv block emphasizes high-frequency energy, pure Mamba block suppresses it, and the HiFi-Mamba block structure achieves a balanced spectrum. This observation motivates our new design that unifies frequency control and dynamic expert routing.

To address these limitations, we propose \textbf{HiFi-MambaV2}, a \emph{hierarchical shared-routed MoE-Mamba} architecture that couples frequency-consistent decomposition with content-adaptive computation, comprising three key components:
\begin{itemize}
    \item \textit{Hierarchical Shared-Routed Mixture-of-Experts (SR-MoE):} performs per-pixel top-1 sparse routing through shared and local experts, enabling adaptive specialization while maintaining cross-depth stability. Multi-level features from different resolutions are concatenated and fed into this module to capture both global and local frequency patterns.
    \item \textit{Separable Frequency-Consistent Laplacian Pyramid (SF-Lap):} employs depthwise binomial filtering and symmetric expansion for energy-preserving, alias-resistant multi-scale decomposition, ensuring stable low- and high-frequency streams.
    \item \textit{Lightweight SE-Guided Residual Path:} injects holistic global context before frequency splitting, serving as an auxiliary enhancement branch within an unrolled, data-consistency-regularized reconstruction backbone.
\end{itemize}

HiFi-MambaV2 improves reconstruction fidelity across diverse anatomies. Section~\ref{sec:experiments} presents consistent gains on benchmark datasets (i.e., {fastMRI}, {CC359}, {ACDC}, {M4Raw}, and {Prostate158}), and validates each component through ablations. The hierarchical shared-routed design establishes a principled balance between global coherence and local detail—reflected in the frequency-domain complementarity visualized in Figure~\ref{fig1}.

\section{Related Work}

\vspace{3pt}\noindent\textbf{Deep Learning-based MRI Reconstruction.}
Early data-driven methods learn a direct mapping from undersampled inputs to artifact-reduced images using CNNs~\cite{jin2017deep, hyun2018deep}. Subsequently, model-based reconstruction methods integrate MR physics by unrolling iterative schemes and inserting data consistency (DC) operators that enforce k-space fidelity. CNNs act as learned priors in image, k-space, or hybrid domains (e.g., ADMM-Net~\cite{sun2016deep}, DuDoRNet\cite{zhou2020dudornet}, ISTA-Net~\cite{zhang2018ista}). These designs improve alias suppression, stability, and interpretability within unrolled iterations. However, convolutional priors are locality-bound and limit long-range anatomical coherence under high acceleration and large fields of view (FOVs).
Moving beyond convolutional priors, vision transformers have been explored to model long-range dependencies and enhance global structural fidelity (e.g., SwinMR~\cite{yun2023swinmr}, ReconFormer~\cite{guo2023reconformer}, FpsFormer~\cite{meng2025boosting}). Self-attention improves context aggregation and boundary continuity. However, the quadratic computational and memory cost impedes scalability to clinical resolutions and large FOVs. In addition, patchwise attention at high resolutions can erode multi-scale contextual coherence~\cite{zhao2024mambamir}.

\vspace{3pt}\noindent\textbf{Mamba-based MRI Reconstruction.}
Mamba is a selective state space model that achieves dynamic long-range dependency modeling through input-dependent gating and selective state transitions. Following its use in vision backbones (e.g., VMamba~\cite{liu2024vmamba}), Mamba has achieved promising results in MRI reconstruction (e.g., MambaMIR~\cite{zhao2024mambamir}, LMO~\cite{li2025lmo}). Despite these advances, existing Mamba-based reconstruction models still face key limitations: redundant multi-directional scanning, local state parameterization weakens anatomical coherence, and a lack of frequency control. Our prior work, HiFi Mamba \cite{chen2025hifi}, addressed part of this gap through dual-stream frequency decoupling. However, it still lacked adaptive routing and global coordination between high- and low-frequency paths.

\vspace{3pt}\noindent\textbf{Mixture-of-Experts and Adaptive Routing.}
The Mixture of Experts (MoE) paradigm \cite{shazeer2017outrageously,lepikhin2021gshard} scales model capacity by routing feature tokens to specialized expert modules, achieving greater expressiveness with sparse computation.
Recent variants incorporate hierarchical routing and load balancing to stabilize training and improve specialization \cite{fedus2022switch,zoph2022st}.
MoE architectures have demonstrated effectiveness across language \cite{du2022glam} and vision domains, enhancing feature diversity and adaptive computation. However, their potential remains underexplored in MRI reconstruction, where adaptive expert routing could be particularly beneficial for balancing global structural reasoning with localized frequency preservation.
Our HiFi-MambaV2 builds on this insight by integrating a hierarchical shared–routed Mixture-of-Experts within a frequency-aware Mamba backbone, enabling content-adaptive dispatch between shared and routed experts specialized for different frequency components. This design allows the model to dynamically balance global coherence and high-frequency detail during reconstruction.

\begin{figure*}[!t]
    \centering
    \includegraphics[width=1.0\linewidth]{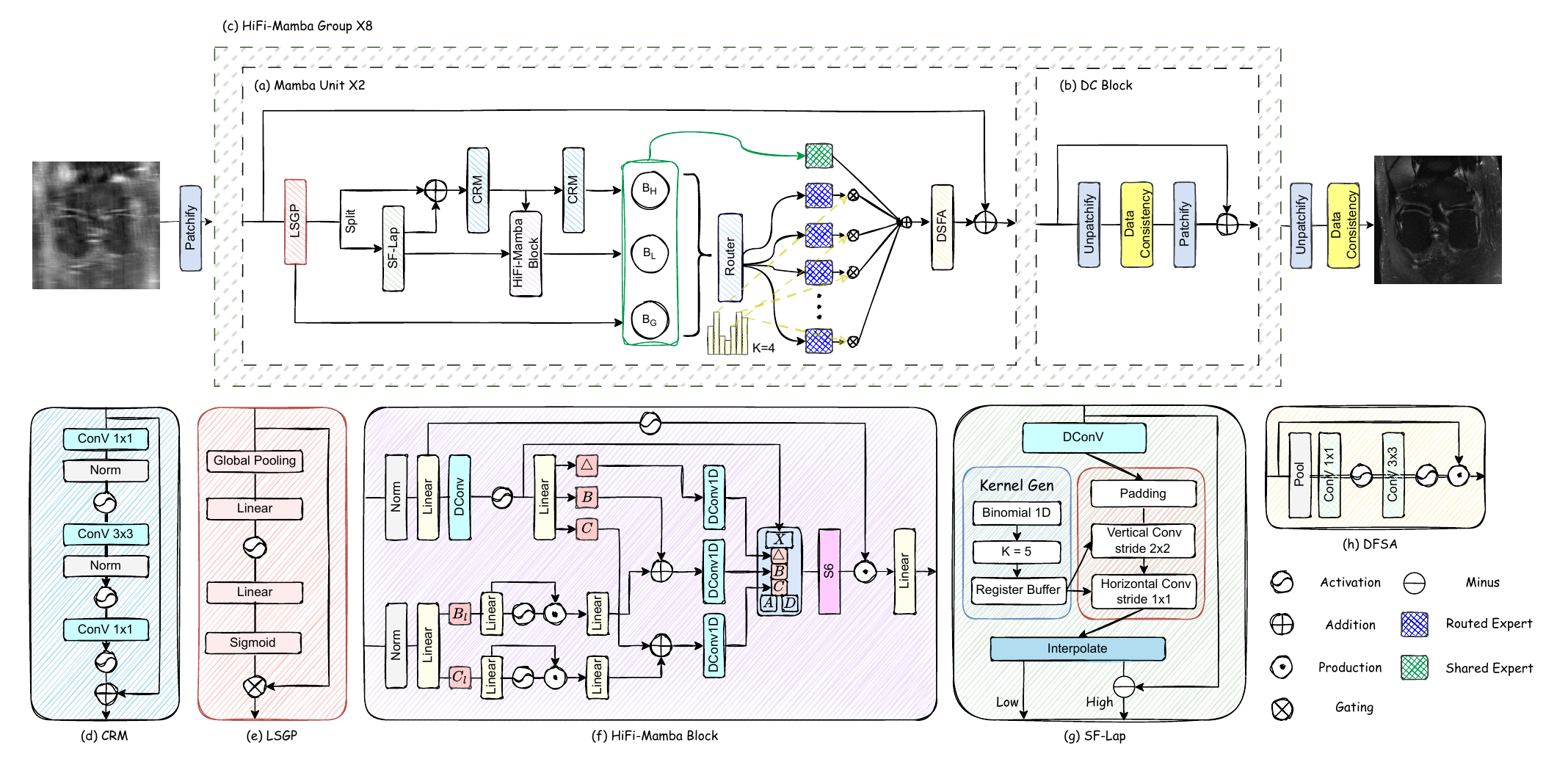}
    \caption{Overview of the proposed HiFi-MambaV2 architecture.
(a) The HiFi-Mamba Unit (×2 per Group) applies the Lightweight SE-Guided Global Context Path (LSGP) module for residual global enhancement, followed by frequency decomposition via the SF-Lap module to produce stable high- and low-frequency components. Each frequency stream is processed by frequency-aware HiFi-Mamba Blocks and routed through a hierarchical shared-routed MoE mechanism with four paths. (b) The Data Consistency (DC) block enforces fidelity with acquired k-space data after each Group.
(c) The HiFi-Mamba Group (×8 cascades) stacks two Mamba Units and one DC block within an unrolled optimization pipeline.
(d) Condition Refinement Module (CRM) performs cross-resolution feature transformation.
(e) The LSGP enhances global anatomical awareness.
(f) The HiFi-Mamba block models frequency-aware sequences using Mamba-based token mixing. 
(g) The Separable Frequency-Consistent Laplacian Pyramid (SF-Lap) achieves energy-preserving, alias-resistant multi-scale decomposition using depthwise binomial filtering and symmetric reconstruction.
(h) The Dual-Frequency Spatial Attention (DFSA) fuses the routed experts’ outputs through adaptive weighting to form the final feature representation.}
    \label{fig:overview}
\end{figure*}
\section{Motivation} \label{sec:motivation}

\vspace{3pt}\noindent\textbf{Global Context Enhancement.}
HiFi-Mamba adopts dual-stream frequency modeling with cross-stream guidance, where high-frequency features condition the low-frequency global context. While this alleviates locality bias, the backbone lacks an independent pathway to consolidate global context before frequency decomposition. We therefore introduce a lightweight Squeeze-and-Excitation (SE)–guided global branch that runs in parallel with the backbone, aggregates holistic context into a stable global representation, and injects this representation into frequency-specific modules and expert routing. This ensures each frequency stream receives coherent context before specialization.

\vspace{3pt}\noindent\textbf{Adaptive Frequency Allocation.}
Routing between high- and low-frequency streams in HiFi-Mamba is fixed across layers. Such a static configuration ignores spatial and anatomical variability in MRI, allocating identical compute to regions with vastly different textural and structural complexity. Consequently, fine anatomical details may be under-modeled, while homogeneous areas are over-provisioned.
HiFi-MambaV2 addresses this limitation with a \textbf{\textit{hierarchical shared–routed MoE}} that performs \textit{per-pixel top-1 sparse routing} to experts. Concretely, \emph{shared experts} are always active and provide a stable, general prior, while \emph{routed experts} are selectively activated by a local (per-layer) token router conditioned on content. Multi-level features from different streams are concatenated and fed into the MoE, enabling adaptive compute allocation between high- and low-frequency experts, encouraging frequency specialization, and maintaining stable behavior across depth. (See the frequency-domain evidence in Fig.~\ref{fig1}.)

\vspace{3pt}\noindent\textbf{Frequency Decomposition.}
The $w$-Laplacian decomposition in HiFi-Mamba relies on fixed, non-learnable filters, which can attenuate global information and introduce aliasing under undersampled MRI. HiFi-MambaV2 replaces this with a separable, frequency-consistent Laplacian pyramid (SF-Lap) that couples depthwise binomial filtering with symmetric padding, preserving signal energy and resisting aliasing. The resulting stable low- and high-frequency streams are then processed by shared and routed experts.

\section{Method} 
\subsection{Overall Architecture}
\label{sec:overall}
Let $\mathbf{x}\!\in\!\mathbb{C}^{H\times W}$ denote the fully-sampled MR image, and 
$\mathbf{y}_c\!\in\!\mathbb{C}^{M}$ the undersampled $k$-space measurements from the $c$-th coil.
Given coil sensitivity maps $\mathbf{S}_c$ and undersampling operator $\mathcal{F}_u$, the multi-coil forward model is
\begin{equation}
\mathbf{y}_c = \mathcal{F}_u(\mathbf{S}_c\mathbf{x}) + \mathbf{n}_c,
\end{equation}
where $\mathbf{n}_c$ denotes measurement noise.
Reconstruction seeks to estimate $\mathbf{x}$ from $\{\mathbf{y}_c\}_{c=1}^{C}$ by jointly enforcing data fidelity and learned regularization.

HiFi-MambaV2 addresses this inverse problem through a hierarchical unrolled framework that integrates frequency-consistent decomposition, global context modeling, and content-adaptive computation within a data-consistency pipeline.
At iteration $t$, the reconstruction update follows
\begin{equation}
\mathbf{x}^{(t+1)}=\mathcal{D}_{\boldsymbol{\theta}}\!\big(\mathbf{x}^{(t)}-\lambda\sum_{c}\mathbf{S}_c^{H}\mathcal{F}_u^{H}(\mathcal{F}_u(\mathbf{S}_c\mathbf{x}^{(t)})-\mathbf{y}_c)\big),
\end{equation}
where $\mathcal{D}_{\boldsymbol{\theta}}$ denotes the learnable feature refinement module.

The overall architecture consists of eight HiFi-Mamba Groups, each corresponding to one unrolled iteration, as illustrated in Fig.~\ref{fig:overview}.
Each group contains two HiFi-Mamba Units followed by one data-consistency (DC) block, forming a reconstruction stage that alternates between feature refinement in the image domain and physics-based consistency in $k$-space.
Each unit contains an SF-Lap that decomposes features into high- and low-frequency streams, a SE-guided Global Context Path that captures global channel dependencies, and a Shared-Routed MoE that adaptively modulates the fused representation, modeling both local and long-range dependencies with linear-time complexity.
This hierarchical design progressively aligns frequency components and enforces anatomical coherence across iterations, yielding reconstructions that preserve both global structure and fine spatial detail.


\subsection{Separable Frequency Laplacian Split (SF\text{-}Lap)}
\label{sec:sflap}
To achieve alias-resistant and frequency-balanced separation of feature tensors, SF-Lap reformulates the classical Gaussian–Laplacian operation into a single-stage, depthwise-separable decomposition that couples (i) binomial low-pass filtering and (ii) bilinear frequency-balanced expansion.

\vspace{3pt}\noindent\textbf{Separable binomial filtering.}
We employ a normalized $5$-tap binomial kernel 
$\mathbf{k}=[1,4,6,4,1]/16$ that approximates a Gaussian with $\sigma\!\approx\!1$ and provides isotropic smoothing.
The kernel is applied horizontally and vertically in a depthwise-separable manner with reflect padding,
reducing the 2-D convolutional cost from $\mathcal{O}(k^2)$ to $\mathcal{O}(2k)$ while preserving Gaussian-like smoothness.

\vspace{3pt}\noindent\textbf{Low\text{-}pass reduction.}
Prior to decimation, a two-pass low-pass with reflect padding suppresses aliasing:
\begin{equation}
\mathbf{L}
= \mathrm{Conv}_{v}^{(s=2)}\!\Big(
    \mathrm{Conv}_{h}^{(s=1)}\big(\mathcal{P}_{\mathrm{ref}}(\mathbf{x}),\,\mathbf{k}\big),\,\mathbf{k}
\Big).
\end{equation}

Here, $\mathrm{Conv}^{(s)}_h(\cdot, k)$ and $\mathrm{Conv}^{(s)}_v(\cdot, k)$ denote depthwise separable 1D convolutions with a horizontal $1\times 5$ binomial kernel and a vertical $5\times 1$ binomial kernel, respectively. We first apply a horizontal pass at unit stride $s=1$ to smooth along the width, and then apply a vertical pass with stride $s=2$ to jointly low-pass filter and perform $2\times$ decimation along both spatial axes. This separable implementation reduces the cost from $O(k^2 HW)$ for a full $k\times k$ 2D convolution to $O(2k HW)$.

\vspace{3pt}\noindent\textbf{High\text{-}frequency extraction and upsampling.}
The Laplacian residual is obtained by interpolating the low-frequency component to the original resolution:
\begin{equation}
    \tilde{\mathbf{L}}=\mathcal{U}_{\times 2}(\mathbf{L}),\qquad
    \mathbf{H}=\mathbf{x}-\tilde{\mathbf{L}},
\end{equation}
where $\mathcal{U}_{\times2}$ denotes $2\times$ bilinear upsampling. Compared with zero-insertion expansion followed by separable filtering, bilinear interpolation offers a smoother synthesis path and avoids checkerboard artifacts in the feature domain.


\subsection{Lightweight SE\text{-}Guided Global Context Path}
\label{sec:globalpath}
Although SF\text{-}Lap enables frequency-aware decomposition, the backbone lacks an explicit mechanism to capture global anatomical context before frequency separation.
To enhance global anatomical awareness, we introduce a lightweight \emph{SE-guided global context path} that performs channel-wise gating based on globally aggregated features.
Given an input feature tensor $\mathbf{x}\!\in\!\mathbb{R}^{B\times H\times W\times C}$, 
global spatial statistics are captured through global average pooling (GAP):
\begin{equation}
\mathbf{z}=\mathrm{GAP}(\mathbf{x})
=\frac{1}{H W}\sum_{i=1}^{H}\sum_{j=1}^{W}\mathbf{x}_{i,j}.
\end{equation}
A compact two-layer gating network generates channel-wise modulation weights:
\begin{equation}
\mathbf{s}=\mathrm{Sigmoid}\!\big(W_2\,\mathrm{ReLU}(W_1\mathbf{z})\big),
\end{equation}
where $W_1\!\in\!\mathbb{R}^{C\times(C/r)}$ and $W_2\!\in\!\mathbb{R}^{(C/r)\times C}$ are linear projections with reduction ratio $r$.
The final output is obtained via residual channel-wise modulation:
\begin{equation}
\mathbf{x}'=\mathbf{x}\odot\mathbf{s}+\mathbf{x},
\end{equation}
where $\odot$ denotes element-wise multiplication with broadcasting along $(H,W)$.
This SE-guided path operates in parallel with the main reconstruction backbone and injects global anatomical cues before the SF-Lap module,
enhancing long-range structural consistency with negligible computational overhead.


\subsection{Hierarchical Shared\text{-}Routed MoE}
\label{sec:moe}
While the SE-guided path enhances global contextual awareness and SF-Lap separates features by frequency,
these modules still operate under a fixed computational pathway that does not adapt to the spatially varying complexity of anatomical structures.
Regions with high textural detail or aliasing artifacts require greater modeling capacity than homogeneous areas.
In MR images, where signal sparsity and anatomical complexity vary spatially, pixel-wise routing enables adaptive frequency modeling across tissue boundaries and aliasing regions.
To this end, we introduce a \emph{hierarchical shared-routed Mixture-of-Experts (MoE)} module that dynamically allocates expert specialization across pixels.

\vspace{3pt}\noindent\textbf{Shared and routed experts.}
Given a fused feature map $\mathbf{x}\!\in\!\mathbb{R}^{B\times H\times W\times C}$ formed by concatenating global, high-, and low-frequency streams,
the module comprises two expert groups: \emph{shared experts}, which are always active for all tokens, and \emph{routed experts}, which are adaptively selected by a lightweight pixel-wise router.
Each expert is implemented as a lightweight channel-wise MLP operating independently across spatial locations.
The outputs of the shared experts are aggregated as
\begin{equation}
\mathbf{y}_{\mathrm{shared}}=\sum_{i=1}^{N_s}E_i^{(\mathrm{sh})}(\mathbf{x}),
\end{equation}
where $N_s$ is the number of shared experts.
A pixel-wise router produces gating logits over $N_r$ routed experts:
\begin{equation}
G=\mathrm{Softmax}\!\big(W_r\,\mathbf{x}\big),\qquad
G\in\mathbb{R}^{B\times H\times W\times N_r},
\end{equation}
where $W_r$ is a $1{\times}1$ linear projection.
Top-1 selection is applied per pixel to obtain the routing mask:
\begin{equation}
\mathbf{M}_{b,h,w,e}= 
\begin{cases}
1,& e=\arg\max G_{b,h,w,:},\\
0,& \text{otherwise.}
\end{cases}
\end{equation}
The routed output is then computed as
\begin{equation}
\mathbf{y}_{\mathrm{routed}}
=\sum_{e=1}^{N_r}\mathbf{M}_{:,:,:,e}\odot E_e^{(\mathrm{rt})}(\mathbf{x}),
\end{equation}
and the final output combines both expert groups:
\begin{equation}
\mathbf{y}=\mathbf{y}_{\mathrm{shared}}+\mathbf{y}_{\mathrm{routed}}.
\end{equation}

The shared-routed MoE module is placed after SF-Lap and the SE-guided path within each reconstruction stage.
Shared experts capture globally consistent responses, while routed experts specialize in localized, frequency-dependent variations.
Since routing is top-1 and all experts share lightweight channel-MLP structures, the computational overhead is negligible.
The hierarchical formulation maintains consistent routing behavior across depths, ensuring stable specialization throughout the unrolled reconstruction.

\vspace{3pt}\noindent\textbf{Load-balancing regularization.}
Pure top-1 routing can lead to expert collapse, where only a small subset of
experts are frequently selected while others remain idle. To encourage more
uniform expert utilization, we import a lightweight load-balancing
regularizer on the router outputs. Let
\begin{equation}
p_e = \frac{1}{BHW} \sum_{b,h,w} G_{b,h,w,e}
\end{equation}
denote the average routing probability for expert $e$. We define
\begin{equation}
\mathcal{L}_\text{bal} = N_r \sum_{e=1}^{N_r} p_e^2,
\end{equation}
which is minimized when all experts receive equal mass $p_e = 1/N_r$. And our 
final loss function could be described as:
\begin{equation}
\mathcal{L} =
\underbrace{\frac{1}{B}\sum_{b=1}^{B}
\big\|\hat{x}_b - x_b^{\mathrm{gt}}\big\|_1}_{\mathcal{L}_{\mathrm{rec}}}
+ \lambda_{\mathrm{bal}}\mathcal{L}_\text{bal}
\label{eq:total-loss}
\end{equation}
where $B$ is the batch size, and $\lambda_{\mathrm{bal}}$ controls
the strength of the load-balancing regularization. $\hat{x}_b \in \mathbb{C}^{H\times W}$ denote the final reconstruction of the
$b$-th sample and $x_b^{\mathrm{gt}}$ its ground truth. 


\section{Experiments}
\label{sec:experiments}

\begin{table*}[t]
\centering
\footnotesize

\begin{tabular}{l|cc|cc|cc|cc|cc|cc}
\hline
\multirow{3}{*}{\textbf{Method}} & \multicolumn{6}{c|}{\textbf{fastMRI}} & \multicolumn{6}{c}{\textbf{CC359}} \\
\cline{2-13}
& \multicolumn{2}{c|}{\textbf{PSNR} $\uparrow$} & \multicolumn{2}{c|}{\textbf{SSIM} $\uparrow$} & \multicolumn{2}{c|}{\textbf{NMSE} $\downarrow$} & \multicolumn{2}{c|}{\textbf{PSNR} $\uparrow$} & \multicolumn{2}{c|}{\textbf{SSIM} $\uparrow$} & \multicolumn{2}{c}{\textbf{NMSE} $\downarrow$} \\
\cline{2-13}
& AF=4 & AF=8 & AF=4 & AF=8 & AF=4 & AF=8 & AF=4 & AF=8 & AF=4 & AF=8 & AF=4 & AF=8 \\
\hline
Zero-Filling & 29.25 & 25.95 & 0.723 & 0.620 & 0.035 & 0.064 & 24.79 & 21.27 & 0.725 & 0.576 & 0.053 & 0.120 \\
\hline
UNet (MICCAI 2015) & 31.66 & 28.60 & 0.798 & 0.697 & 0.021 & 0.035 & 28.27 & 24.28 & 0.847 & 0.720 & 0.025 & 0.059 \\
ISTA-NET (CVPR 2018) & 33.27 & 29.44 & 0.832 & 0.714 & 0.012 & 0.030 & 32.03 & 25.44 & 0.902 & 0.744 & 0.010 & 0.046 \\
\hline
ReconFormer (TMI 2023) & 33.75 & 30.42 & 0.837 & 0.728 & 0.011 & 0.026 & 32.46 & 26.47 & 0.906 & 0.766 & 0.010 & 0.039 \\
FpsFormer (AAAI 2025) & 33.74 & 30.63 & 0.841 & 0.732 & 0.011 & 0.026 & 32.35 & 26.65 & 0.897 & 0.768 & 0.010 & 0.038 \\
\hline
LMO (CVPR 2025)& 34.49 & 31.10 & 0.846 & 0.744 & 0.011 & 0.023 & 35.35 & 27.99 & 0.921 & 0.787 & 0.006 & 0.028\\
HiFi-MambaV1(P2) (AAAI 2026)& 34.47 & 31.38 & 0.853 & 0.758 & 0.010 & 0.021 & 35.74 & 28.08 & 0.935 & 0.802 & 0.005 & 0.027\\
HiFi-MambaV1(P1) (AAAI 2026)& 34.85 & 31.81 & 0.855 & 0.762 &  0.010 & 0.020 & 36.93 & 28.49 & 0.942 & 0.810 & 0.004 & 0.026 \\
HiFi-MambaV2(P2) & 34.63 & 31.73 & 0.854 & 0.763 & 0.010 & 0.019 & 36.80 & 28.68 & 0.955 & 0.841 & 0.003 & 0.022\\
\textbf{HiFi-MambaV2(P1)} & \textbf{34.86} & \textbf{31.89} & \textbf{0.855} & \textbf{0.764} &  \textbf{0.010} & \textbf{0.019} & \textbf{37.21} & \textbf{29.05} & \textbf{0.957} & \textbf{0.849} & \textbf{0.003} & \textbf{0.020} \\

\hline
\end{tabular}
\caption{Quantitative comparison on the fastMRI-Equispaced and CC359-Equispaced under $4\times$ and $8\times$ acceleration factors.}
\label{tab:mri_comparison}
\end{table*}

\subsection{Experimental Settings}
\vspace{3pt}\noindent\textbf{Datasets.}
We evaluate HiFi-MambaV2 on five publicly available MRI datasets: fastMRI (knee)~\cite{zbontar2018fastmri}, CC359 (brain)~\cite{warfield2004staple}, ACDC (cardiac)~\cite{bernard2018deep}, M4Raw (multi-coil brain)~\cite{lyu2023m4raw}, and Prostate158 (prostate)~\cite{adams2022prostate158}. 
I) \textbf{fastMRI.} The dataset comprises 1,172 single-coil coronal knee scans with Proton Density Fat-Suppressed (PDFS) weighting. We follow the official train/validation/test split.  
II) \textbf{CC359.} Raw brain MR volumes acquired, with 25 subjects for training and 10 for testing. 
III) \textbf{ACDC.} Cine cardiac MR images of 150 subjects. Following the official split, 100 cases are used for training and 50 for testing
IV) \textbf{M4Raw.} This dataset includes multi-contrast, multi-repetition, and multi-channel brain MRI k-space data from 183 volunteers scanned at 0.3T with a four-channel coil. 1,024 and 240 volumes are used for training and validation, respectively. 
V) \textbf{Prostate158.} The dataset provides prostate MR images from 158 volunteers, split into 139 training and 19 testing cases.
To ensure consistent anatomical content, the first and last five slices of fastMRI and the first and last 15 slices of CC359 are discarded. Data preprocessing is shown in Appendix A.1. Data introduction is described in Appendix A.2.

\vspace{3pt}\noindent\textbf{Evaluation Metrics.}
Reconstruction quality is evaluated using three standard metrics: Peak Signal-to-Noise Ratio (PSNR)~\cite{huynh2008scope}, Structural Similarity Index (SSIM)~\cite{wang2004image}, and Normalized Mean Squared Error (NMSE)~\cite{zhao2016loss}.

\begin{figure*}[htbp]
    \centering
    \includegraphics[width=\linewidth]{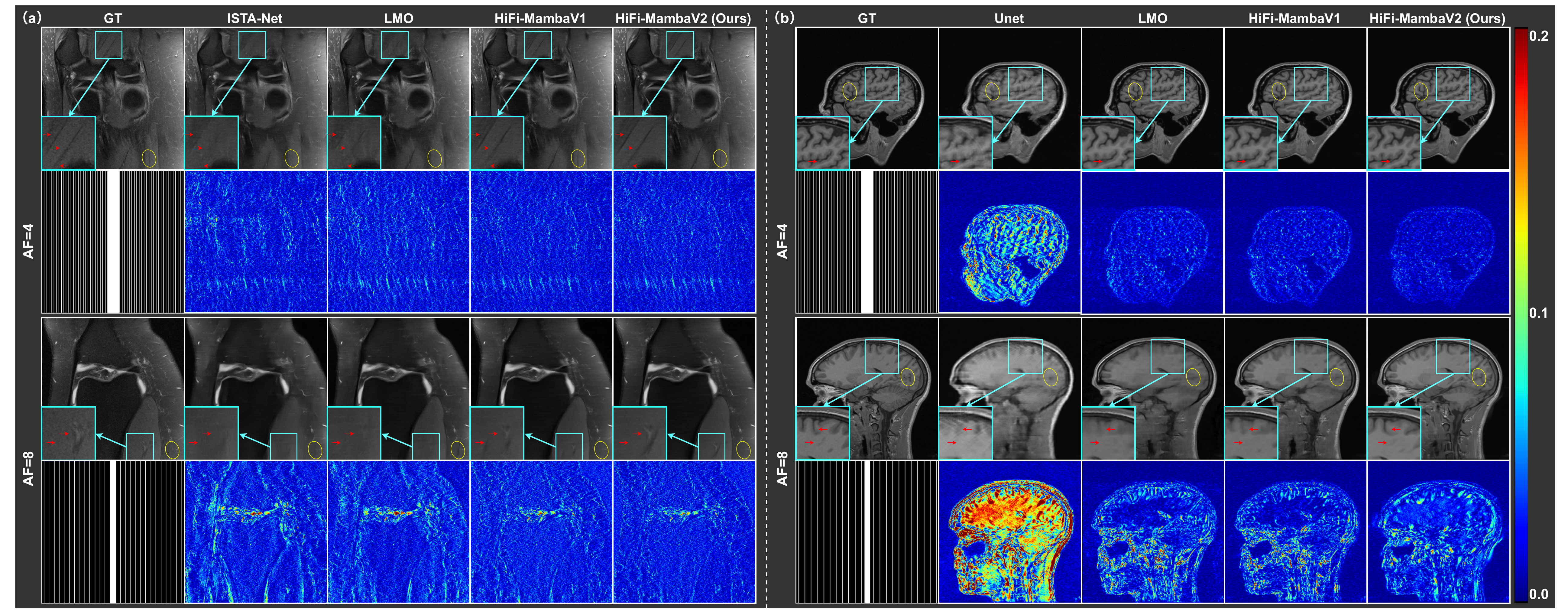}
    \caption{Qualitative comparison on the fastMRI and CC359 datasets under single-coil settings. (a) Reconstruction results on the fastMRI knee dataset with acceleration factors AF=4 and AF=8. (b) Reconstruction results on the CC359 brain dataset under the same acceleration factors. The second row of each subplot shows the corresponding error maps. The blue boxes, yellow ellipses and red arrow highlight the details in the reconstruction results.}
    \label{fig:result}
\end{figure*}
\vspace{3pt}\noindent\textbf{Training Details.}
Training employs the AdamW optimizer with an initial learning rate of $8\times10^{-4}$, cosine annealing schedule, and a five-epoch warm-up. The model is trained for 100 epochs. All experiments are conducted on two NVIDIA H100 GPUs; FLOPs are measured on an NVIDIA A100.

\begin{table}[t]
\footnotesize
\centering
\begin{tabular}{c|l|ccc}
\hline
\textbf{AF} & \textbf{Methods} & \textbf{PSNR} & \textbf{SSIM} & \textbf{NMSE} \\
\hline
\multirow{5}{*}{$\times 4$}
& ISTA-NET        & 27.23 & 0.784 & 0.017 \\
& LMO      & 28.15 & 0.808 & 0.015 \\
& HiFi-MambaV1(P1)     & 28.85 & 0.833 & 0.013 \\
& HiFi-MambaV2(P2)       &28.88 & 0.832 & 0.013 \\
& \textbf{HiFi-MambaV2(P1)} & \textbf{29.04} & \textbf{0.837} & \textbf{0.012} \\
\hline
\multirow{5}{*}{$\times 8$}
& ISTA-NET        & 22.02 & 0.579 & 0.055 \\
& LMO      & 22.68 & 0.600 & 0.052 \\
& HiFi-MambaV1(P1)     & 23.22 & 0.654 & 0.046 \\
& HiFi-MambaV2(P2)       & 23.61& 0.656 & 0.038 \\
& \textbf{HiFi-MambaV2(P1)} & \textbf{23.91} & \textbf{0.662} & \textbf{0.039} \\
\hline
\end{tabular}
\caption{Single-coil MRI reconstruction comparison at acceleration factor $\times 4$ and $\times 8$ with random mask at Prostate158.}
\label{tab:prostate_random}
\end{table}

\subsection{Comparison with State-of-the-Art Methods}
\subsubsection{Quantitative Results}

Tables~\ref{tab:mri_comparison}--\ref{tab:m4raw_multicoil} present comprehensive quantitative comparisons across multiple datasets, sampling patterns, and acceleration factors.  
HiFi-MambaV2 consistently achieves the best or second-best performance in all settings, demonstrating stable reconstruction accuracy and strong generalization across both single- and multi-coil configurations.

\vspace{3pt}\noindent\textbf{Equispaced sampling (fastMRI and CC359).}
As summarized in Table~\ref{tab:mri_comparison}, HiFi-MambaV2 delivers consistent gains over CNN-, Transformer-, and previous Mamba-based architectures under both $4\times$ and $8\times$ acceleration.  
On fastMRI, V2(P1) attains 34.86\,dB/0.855 SSIM at $4\times$ and 31.89\,dB/0.764 SSIM at $8\times$, outperforming LMO by +0.37\,dB/+0.011 and +0.79\,dB/+0.020, respectively.  
On CC359, V2(P1) achieves 37.21\,dB/0.957 SSIM at $4\times$ and 29.05\,dB/0.849 SSIM at $8\times$, surpassing LMO by +1.86\,dB/+0.036 and +1.06\,dB/+0.062.  
The corresponding NMSE reductions confirm that HiFi-MambaV2 achieves the highest fidelity under both moderate and heavy undersampling.

\begin{table}[t]
\footnotesize
\centering
\begin{tabular}{c|l|ccc}
\hline
\textbf{AR} & \textbf{Methods} & \textbf{PSNR} & \textbf{SSIM} & \textbf{NMSE} \\
\hline
\multirow{5}{*}{$\times 4$}
& Unet        & 25.19 & 0.796 &0.033 \\
& LMO      & 28.74 & 0.833 & 0.018 \\
& HiFi-MambaV1(P1)     & 33.48& 0.928 & 0.005 \\
& HiFi-MambaV2(P2)       & 32.89& 0.923 & 0.005 \\
& \textbf{HiFi-MambaV2(P1)} & \textbf{33.72} & \textbf{0.931} & \textbf{0.005} \\
\hline
\multirow{5}{*}{$\times 8$}
& Unet        & 22.21 & 0.640 & 0.065 \\
& LMO      & 24.18 & 0.677 & 0.043 \\
& HiFi-MambaV1(P1)     & 27.64&  0.805& 0.018 \\
& HiFi-MambaV2(P2)       & 27.44 & 0.804 & 0.019 \\
& \textbf{HiFi-MambaV2(P1)} & \textbf{28.05} & \textbf{0.820} & \textbf{0.017} \\
\hline
\end{tabular}
\caption{Single-coil MRI reconstruction comparison at acceleration rate $\times 4$ and $\times 8$. under radial undersampling mask at ACDC.}
\label{tab:acdc_radial}
\end{table}

\begin{table}[t]
\footnotesize
\centering
\begin{tabular}{c|l|ccc}
\hline
\multirow{5}{*}{$\times 4$}
& Unet        & 30.02 & 0.770 & 0.026 \\
& LMO      & 29.86 & 0.769  & 0.026 \\
& HiFi-MambaV1(P1)     & 31.81 &  0.794 &  0.016\\
& HiFi-MambaV2(P2)       & 31.27 & 0.781 & 0.018 \\
& \textbf{HiFi-MambaV2(P1)} & \textbf{31.82} & \textbf{0.794} & \textbf{0.016} \\
\hline
\multirow{5}{*}{$\times 8$}
& Unet        & 28.14 & 0.724 & 0.037 \\
& LMO      & 28.35 & 0.728 & 0.036 \\
& HiFi-MambaV1(P1)     & 29.89 & 0.746 & 0.025 \\
& HiFi-MambaV2(P2)       & \textbf{29.91} & \textbf{0.747} & \textbf{0.024}  \\
& \textbf{HiFi-MambaV2(P1)} & 29.83 & 0.744 & 0.026 \\
\hline
\end{tabular}
\caption{Multi-coil MRI reconstruction comparison at acceleration rate  $\times 4$ and $\times 8$. at M4Raw.}
\label{tab:m4raw_multicoil}
\end{table}

\vspace{3pt}\noindent\textbf{Random mask (Prostate158, single-coil).}
Under random undersampling (Table~\ref{tab:prostate_random}), HiFi-MambaV2 again leads across all metrics.  
At $4\times$, V2(P1) achieves 29.04\,dB PSNR and 0.837 SSIM, improving upon LMO by +0.89\,dB/+0.029 with a lower NMSE (0.012 vs.\ 0.015).  
At $8\times$, the gap further widens to +1.23\,dB/+0.062 and a 25\% NMSE reduction, demonstrating the robustness of the proposed model to non-uniform sampling patterns.

\vspace{3pt}\noindent\textbf{Radial mask (ACDC, single-coil).}
Table~\ref{tab:acdc_radial} shows that HiFi-MambaV2(P1) maintains top performance under radial sampling.  
At $4\times$, V2(P1) records 33.72\,dB/0.931 SSIM, slightly exceeding V1(P1) by +0.24\,dB/+0.003.  
At $8\times$, V2(P1) reaches 28.05\,dB/0.820 SSIM, outperforming LMO by +3.87\,dB/+0.143 while matching or exceeding prior HiFi-MambaV1 performance.  
These results validate the model’s stability under non-Cartesian trajectories and high acceleration.

\vspace{3pt}\noindent\textbf{Multi-coil reconstruction (M4Raw).}
On the M4Raw multi-coil dataset (Table~\ref{tab:m4raw_multicoil}), HiFi-MambaV2 achieves comparable or superior performance to all baselines.  
At $4\times$, both V1(P1) and V2(P1) achieve the highest PSNR (31.82\,dB) and SSIM (0.794), while V2(P2) attains the best efficiency–accuracy trade-off.  
At $8\times$, V2(P2) achieves the highest PSNR (29.91\,dB) and lowest NMSE (0.024), outperforming LMO by +1.56\,dB and a 33\% NMSE reduction.  
Across all acceleration factors, HiFi-MambaV2 demonstrates strong generalization to multi-channel inputs and consistent numerical superiority.

Overall, quantitative comparisons across Tables~\ref{tab:mri_comparison}–\ref{tab:m4raw_multicoil} confirm that HiFi-MambaV2 achieves state-of-the-art reconstruction accuracy with consistently lower NMSE across diverse sampling schemes and datasets, validating its scalability and robustness.


\subsubsection{Qualitative Results}
Figure~\ref{fig:result} compares reconstruction results at $4\times$ and $8\times$ acceleration on the fastMRI (knee) and CC359 (brain) datasets. 
Each case shows the reconstructed image (top) and the corresponding error map (bottom), color-coded from 0 (blue) to 0.2 (red). 
Yellow circles and red arrows highlight structural discrepancies, and blue boxes mark zoomed-in regions.

CNN-based baselines (ISTA-Net, UNet) exhibit edge blurring and loss of fine detail, while LMO, despite using the Mamba backbone, still shows boundary degradation and missing textures. 
Both HiFi-Mamba variants recover clearer structures, and HiFi-MambaV2 further enhances tissue sharpness and smooth intensity transitions, particularly at $8\times$ acceleration. 
Error maps show that HiFi-MambaV2 produces lower and more uniform residuals, effectively reducing aliasing and streaking artifacts.
Overall, HiFi-MambaV2 achieves the most accurate and stable reconstructions across anatomies and acceleration factors.

\subsubsection{Efficiency Analysis}
We evaluate the efficiency of HiFi-MambaV2 on fastMRI at $8\times$ acceleration with $320\times320$ resolution. Compared with Transformer-based baselines such as ReconFormer and FpsFormer, HiFi-MambaV2 remains substantially more efficient while delivering higher reconstruction accuracy. Although the hierarchical routing and SF-Lap modules slightly increase theoretical FLOPs compared to HiFi-MambaV1, they introduce a finer frequency decomposition and adaptive token dispatch that significantly enhance reconstruction stability and generalization across datasets. The heavier variant HiFi-MambaV2(P1) achieves the highest reconstruction quality (31.89\,dB PSNR, 0.764 SSIM), surpassing all competing models including LMO (484.98\,G FLOPs) by +0.79\,dB/+0.020 SSIM. Importantly, the lighter configuration HiFi-MambaV2(P2) attains nearly identical performance (31.73\,dB/0.763 SSIM) with only 172.21\,G FLOPs—approximately one-third of LMO and comparable to previous HiFi-Mamba(P1). This demonstrates that HiFi-MambaV2 scales gracefully across computational budgets, offering both high-accuracy and resource-efficient configurations. It is worth noting that the measured wall-clock latency increase is significantly smaller than the theoretical FLOPs growth, since the sparse top-1 routing and depthwise separable operations in SF-Lap are highly parallelized on modern GPUs. Therefore, the additional computation primarily reflects enhanced frequency modeling rather than inefficient computation. Overall, HiFi-MambaV2 achieves superior accuracy–efficiency balance by trading a modest increase in theoretical cost for markedly improved frequency fidelity, robustness, and scalability.

\begin{table}[t]
\centering
\footnotesize

\begin{tabular}{l|c|c|c|c}
\hline
\textbf{Method} & \textbf{FLOPs}  & \textbf{PSNR} & \textbf{SSIM} & \textbf{NMSE}\\
\hline
ReconFormer  & 270.60G & 30.42 & 0.728 & 0.026\\
FpsFormer  & 200.45G & 30.63 & 0.732 & 0.026\\
LMO &   484.98G & 31.10 & 0.744  & 0.023\\
HiFi-Mamba (P1)  & 270.37G &  31.81 & 0.762& 0.021 \\
HiFi-Mamba (P2)  & 67.87G & 31.38 & 0.758 & 0.020\\
HiFi-MambaV2 (P1)  & 687.15G &   31.89 & 0.764  & 0.019\\
\textbf{HiFi-MambaV2 (P2)} & 172.21G &  31.73 & 0.763& 0.019\\
\hline
\end{tabular}
\caption{Efficiency test on image size 320$\times$320 on NVIDIA A100.}
\label{tab:efficiency_cc359}
\end{table}

\subsection{Ablation Study and Analysis}
\label{sec:ablation}

\begin{table}[t]
\footnotesize
\centering

\resizebox{\linewidth}{!}{
\begin{tabular}{cccccccc}
\hline
\textbf{HFM} & \textbf{SF-Lap.} & \textbf{LSGP} & \textbf{MoE-B} & \textbf{MoE} & \textbf{PSNR} & \textbf{SSIM} & \textbf{NMSE} \\
\hline
\checkmark &            &            &            &            & 28.08 & 0.802 & 0.027 \\
\checkmark & \checkmark &            &            &            & 28.31 & 0.830 & 0.024 \\
\checkmark & \checkmark & \checkmark &            &            & 28.43 & 0.835 & 0.023 \\
\checkmark & \checkmark & \checkmark &            & \checkmark & 28.50 & 0.837 & 0.023 \\
\checkmark & \checkmark & \checkmark & \checkmark &            & \textbf{28.68} & \textbf{0.841} & \textbf{0.022} \\
\hline
\end{tabular}} 

\caption{Model component experiment is conducted on the CC359 dataset with AF $=8$, and patch size $=2$. HFM refers to our HiFi-Mamba architecture while MoE-B refers to training with balanced loss.}
\label{tab:cc359_ablation}
\end{table}

\begin{table}[t]
\footnotesize
\centering

\begin{tabular}{ccccc}
\hline
\textbf{Patch Size} & \textbf{Depth} & \textbf{PSNR} $\uparrow$ & \textbf{SSIM} $\uparrow$ & \textbf{NMSE} $\downarrow$ \\
\hline
2 & 3$\times$2 & 27.80 & 0.798 & 0.029 \\
2 & 4$\times$2 & 27.93 & 0.800 & 0.028 \\
2 & 6$\times$2 & 28.36 & 0.823 & 0.024 \\
2 & 8$\times$2 & 28.68 & 0.841 & 0.022 \\
4 & 8$\times$2 & 27.70 & 0.794 & 0.029 \\
1 & 8$\times$2 & \textbf{29.05} & \textbf{0.849} & \textbf{0.020} \\
\hline
\end{tabular}
\caption{
Ablation study on model depth and patch size using the CC359 dataset (AF = 8). }
\label{tab:cc359_ablation_size}
\end{table}

\vspace{3pt}\noindent\textbf{Component-wise Ablation.}
We conduct a component-wise ablation on the CC359 dataset at $8\times$ acceleration to assess the contribution of each module in HiFi-MambaV2. 
As shown in Table~\ref{tab:cc359_ablation}, the baseline HiFi-Mamba (HFM) achieves 28.08\,dB PSNR, 0.802 SSIM, and 0.027 NMSE. 
Adding SF-Lap improves performance to 28.31\,dB and 0.830 SSIM by providing more effective frequency decomposition, and introducing the LSGP further boosts accuracy to 28.43\,dB/0.835 SSIM. Incorporating the hierarchical MoE routing mechanism enables adaptive token allocation across frequency experts, reaching 28.68\,dB PSNR, 0.841 SSIM, and 0.022 NMSE, while training this routing with the balanced loss (MoE-B) stabilizes expert utilization and yields 28.50\,dB/0.837 SSIM. These trends show that SF-Lap, LSGP, and MoE-B provide complementary gains in frequency modeling and global context, whose combination leads to the highest overall fidelity and stability under high acceleration.

\vspace{3pt}\noindent\textbf{Ablation on Model Depth and Patch Size.}
We analyze the impact of model depth and patch size under an $8\times$ acceleration setting on the CC359 dataset. 
As shown in Table~\ref{tab:cc359_ablation_size}, when the patch size is fixed at 2, increasing the depth from $3\times2$ to $8\times2$ yields steady improvements, with PSNR rising from 27.80\,dB to 28.28\,dB and SSIM from 0.798 to 0.815, while NMSE decreases from 0.029 to 0.026. 
This indicates that deeper cascades enhance frequency modeling and anatomical detail, although the gain gradually saturates beyond moderate depth.

With depth fixed at $8\times2$, reducing the patch size from 4 to 1 consistently improves all metrics. 
The finest configuration (patch = 1) achieves 28.68\,dB PSNR, 0.841 SSIM, and 0.022 NMSE, outperforming coarser settings by a clear margin. 
These results confirm that increased model depth and finer patch granularity act synergistically to improve reconstruction fidelity without overfitting, demonstrating the scalability and robustness of HiFi-MambaV2.

\section{Conclusion}

We presented HiFi-MambaV2, a hierarchical shared-routed Mamba architecture for high-fidelity MRI reconstruction. The proposed model unifies frequency decomposition and adaptive computation through two complementary components: a separable frequency Laplacian (SF-Lap) and a hierarchical Mixture-of-Experts with shared experts and local routers, which performs top-1 sparse token dispatch across frequency experts. A lightweight global-context path further enhances long-range dependency modeling while preserving anatomical fidelity. Comprehensive experiments on five public datasets demonstrate that HiFi-MambaV2 consistently surpasses CNN-, Transformer-, and previous Mamba-based approaches in PSNR, SSIM, and NMSE across both single- and multi-coil settings and multiple acceleration factors. Ablation studies confirm the complementary effects of frequency decomposition, hierarchical MoE, and global enhancement, validating the scalability and robustness of the proposed design. Future work will focus on further improving the model’s efficiency and extending its frequency-aware design toward broader clinical scenarios.

\bibliographystyle{IEEEtran}
\bibliography{main}

\clearpage
\newpage
\onecolumn
\section*{Appendix}

\subsection*{A.1 Data Preprocessing}
The preprocessing for fastMRI and CC359 follows the data preprocessing pipeline of HiFi-Mamba. For Prostate158, we adopt the same procedure, except that the sampling mask is replaced with a random mask. For ACDC, we employ a golden-angle radial sampling mask and first perform a center crop to an image size of $128\times128$. Four-fold undersampling corresponds to using 32 radial spokes, while eight-fold undersampling uses 16 spokes. For M4Raw, we use the equispaced Cartesian undersampling mask and split the real and imaginary parts of the four-channel k-space input to form an eight-channel input.

\subsection*{A.2 Dataset Introduction}
The fastMRI dataset used in our study consists of 1,172 complex-valued single-coil coronal knee examinations acquired with proton-density (PD) weighting. Each volume contains approximately 35 coronal slices with an in-plane matrix size of $320\times320$. We train and evaluate all models on the proton-density fat-suppressed (PDFS) subset, strictly following the official train/validation/test split.\\
The CC359 dataset contains raw brain MR scans acquired on clinical MR scanners. Following the official split, we use 25 subjects for training and 10 subjects for testing. Each 2D slice has an in-plane matrix size of $256\times256$.\\
The ACDC dataset consists of cardiac MR images from 150 subjects. Following the official split, we use 100 cases for training and 50 cases for testing, each containing short-axis slices with an in-plane resolution of $256\times256$.\\
The M4Raw dataset consists of multi-channel brain MRI k-space data from 183 volunteers scanned at 0.3T using a four-channel coil. We use 139 volumes for training and 19 volumes for validation.\\
Prostate158 dataset provides prostate MR images from 158 volunteers, split into 139 training and 19 testing cases.\\

\end{document}